\documentclass[10pt,twocolumn,letterpaper]{article}

\usepackage{cvpr}
\usepackage{times}
\usepackage[pdftex]{graphicx}
\usepackage{amsmath}
\usepackage{amssymb}
\usepackage{bm}
\usepackage{multirow,tabularx}
\usepackage{hyperref}
\usepackage{colortbl}
\usepackage{booktabs}
\usepackage{multirow,tabularx}

\begin{document}

\title{A Human-Centered Risk Evaluation of Biometric Systems\\ Using Conjoint Analysis}

\author{
Tetsushi Ohki$^{1,2}$,
Narishige Abe$^{3}$,
Hidetsugu Uchida$^{3}$,
Shigefumi Yamada$^{3}$\\
{$^1$Shizuoka University, Shizuoka, JP}, 
{$^2$RIKEN AIP, Tokyo, JP},
{$^3$Fujitsu Limited, Kawasaki, JP}\\
{\tt\small ohki@inf.shizuoka.ac.jp}, 
{\tt\small \{abe.narishige,u.hidetsugu,yamada.shige\}@fujitsu.com}
}


\maketitle
\thispagestyle{empty}

\begin{abstract}

Biometric recognition systems, known for their convenience, are widely adopted across various fields. However, their security faces risks depending on the authentication algorithm and deployment environment. Current risk assessment methods faces significant challenges in incorporating the crucial factor of attacker's motivation, leading to incomplete evaluations. This paper presents a novel human-centered risk evaluation framework using conjoint analysis to quantify the impact of risk factors, such as surveillance cameras, on attacker's motivation. Our framework calculates risk values incorporating the False Acceptance Rate (FAR) and attack probability, allowing comprehensive comparisons across use cases. A survey of 600 Japanese participants demonstrates our method's effectiveness, showing how security measures influence attacker's motivation. This approach helps decision-makers customize biometric systems to enhance security while maintaining usability.

\end{abstract}


\section{Introduction}
%
Biometric recognition technology recognizes users based on physical or behavioral characteristics, and due to its high convenience, it has been increasingly introduced into a wide range of fields in recent years. The security of biometric recognition systems has been extensively studied, focusing on aspects such as False Acceptance Rate (FAR), spoofing detection \cite{Micheletto2022}, and template protection \cite{Ratha2001}.
%
%
However, when deploying biometric recognition systems in real-world environments, the risk associated with these systems is influenced not only by the technical aspects of the algorithms but also by the presence or absence of security measures in the deployment environment (e.g., surveillance cameras, security personnel). Therefore, a comprehensive risk evaluation that considers these factors is essential.

Numerous studies have addressed the risk assessment of biometric recognition systems. For example, Adler et al. explored attacks on biometric processes that are unrelated to spoofing \cite{Adler2007}. 
Additionally, studies by Dimitriadis \cite{Dimitriadis2004}, Montecchi \cite{Montecchi2012}, Shawn \cite{Shawn2018} and Lai \cite{Lai2023fairness, Lai2023} have focused on organizing and integrating biometric system risks to aid implementers in decision-making.
These studies focus on organizing biometric system risks and integrating those risk values to support decision-making by implementers.
A common approach is to estimate the integrated risk as the product of the occurrence probability of attacks and their impact. However, in practical scenarios, the occurrence probability depends on the attacker's motivation. 
The attacker’s motivation is affected not only by the system's recognition performance but also by various risk factors in the biometric environment, such as the presence of surveillance cameras. Therefore, a human-centered framework is crucial for accurate risk estimation.



This paper proposes a novel human-centered risk evaluation framework that considers the impact of various risk factors on the attacker's motivation. 
This framework allows for the calculation of risk values in practical scenarios, considering both the False Acceptance Rate (FAR) and the probability of attack occurrence, enabling comparison across different use cases.
Our framework adapts conjoint analysis to quantify how various risk factors influence the probability of attack occurrence. In the field of economics, conjoint analysis is a well-known survey method for evaluating the impact of changes on consumer choices. We apply this method to treat factors that influence the attackers motivation to attack the system, as risk factors and quantify their impact on the probability of attack occurrence.
Once the probability of attack occurrence is quantified based on the risk factors, the risk value can be calculated using the risk factors, the False Positive Identification Rate (FPIR) of the authentication algorithm, and the cost of damage due to false acceptance (Figure 1). Subsequently, we compare the risk values of biometric systems in different use cases and demonstrate that users can configure a biometric system that meets their specific risk requirements. 

The contributions of this paper are summarized as follows.

\begin{enumerate}
    \item We propose a novel human-centered risk evaluation framework for biometric systems.
    \item We apply conjoint analysis to quantify the impact of various risk factors on attacker motivation.
    \item A survey of 600 Japanaese participants was conducted to demonstrate the effectiveness of the proposed method. The survey results provides concrete numerical evidence of the effectiveness of different security measures.
\end{enumerate}
\begin{figure}[t]
    \begin{center}
        \includegraphics[width=85mm]{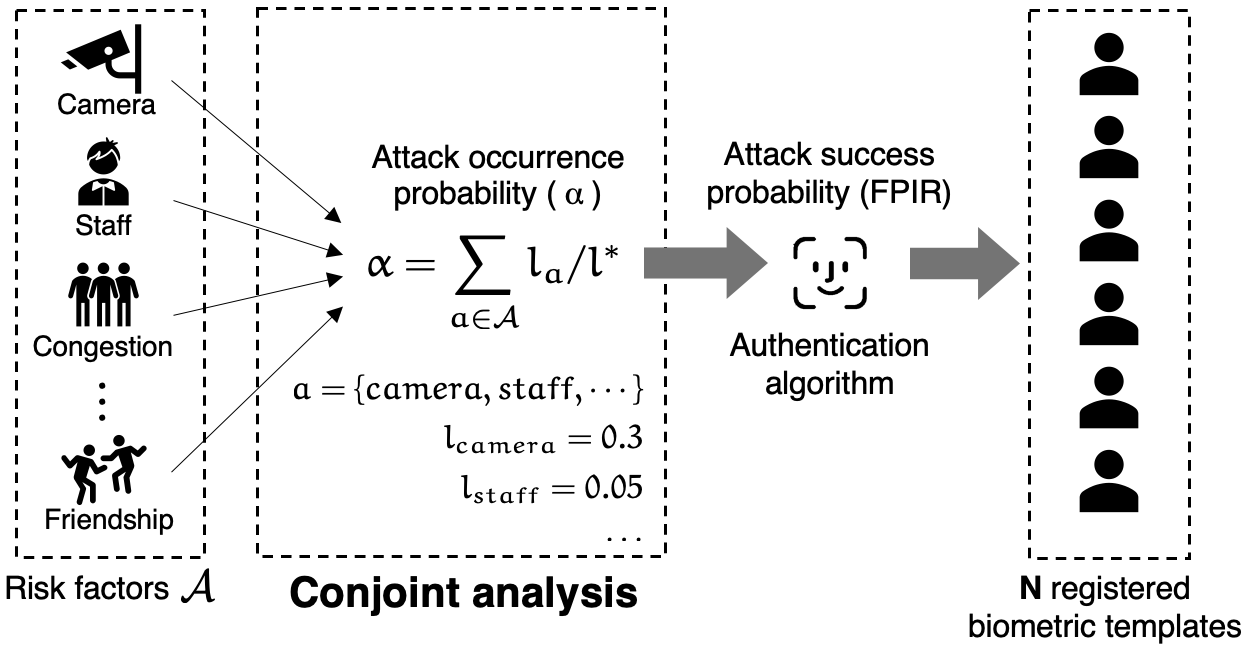}
        \caption{Overview of our proposal}
        \label{fig:paper_overview}
    \end{center}
\end{figure}
\section{Related works}
\subsection{Risk evaluation of Biometrics}
More research reports have been made on the risk evaluation of biometric authentication systems. 

For example, Blanco-Gonzalo et al.\cite{Ramon2019} summarized a user perspective UX assessment of biometric systems. Kohler et al.\cite{Kohler2021} reported the results of a comparative evaluation of biometric authentication system as an alternative to password from reliability, security and usability perspectives.
Eastwood et al.\cite{Shawn2018} examined the risk assessment technique of the biometrics based on the Technology Gap Theory. Lai et al.\cite{Lai2020} offered a complete taxonomy of the R-T-B (risk, trust, and bias) causal performance regulators for the biometric-enabled DSS (decision support systems). Montecchi et al.\cite{Montecchi2012} modeled the threats to biometric authentication systems considering human factors and performed a quantitative security evaluation of the multi-biometric using that model.%
In contrast, various studies have been conducted from a security perspective\cite{Patrizio2013}. In addition to the analysis of vulnerability to threats in general systems, a unique threats in biometric authentication is the presentation attack. Recently, Purnapatra et al.\cite{Purnapatra2021} have proposed many methods, and held a competition (LivDet-Face) at IJCB 2021. Ming et al.\cite{Zuheng2020}  compiled a survey paper on their PAD technologies for facial recognition using a common camera.

These methods assume that the values required for risk assessment can be set observably or arbitrarily. However, predicting how the risk factors introduced in practical system will affect the risk value in often difficult before their introduction. In this paper, we use conjoint analysis to quantify a priori how risk factors to be introduced reduce risk factors and attacker motivation, and incorporate it to perform a practical risk assessment.

\subsection{ISO/IEC 19795-1:2021}
ISO/IEC 19795 series standards were developed by ISO/IEC JTC/1 SC37 for testing and reporting the performance of biometrics systems. In Part 1 of the 19795, ISO/IEC 19795-1\cite{ISO19795}, defines the general principles for testing the performance, including performance metrics.

In particular, Chapter 9 of Part 1 provides various performance evaluation metrics for performance evaluation. It includes the false match rate (FMR) and false non-match rate (FNMR), which evaluate the one-to-one comparison performance, as well as the false acceptance rate (FAR) and false reject rate (FRR), which evaluate the performance of the verification system.
Note that FAR and FRR include the probability of the failure of biometric information acquisition by sensors, given as failure to acquire rate (FTAR).


In addition to these one-to-one verification performance metrics, Section 9.6 includes false negative identification rate (FNIR) and false positive identification rate (FPIR), which evaluate one-to-many identification performance. Since we aim to analyze impersonation risks in this study, particularly for one-to-many identification systems, evaluating how FPIR/FNIR varies depending on risk factors is important. Risk factors consist of security measures such as the presence of surveillance cameras.




\subsection{NIST SRE}
The NIST SRE\cite{NISTSRE} is a large-scale contest to evaluate speaker recognition performance. It was held annually from 1996 to 2006 and every other year since then. Two performance evaluation metrics are used in general speaker recognition tasks, including $P_{Miss}(\theta)$, which is the probability of misidentifying a target user as a non-target user with threshold $\theta$, and $P_{FalseAlarm}(\theta)$, the probability of misidentifying a non-target user as a target user with threshold $\theta$.
Because FRR and FAR have a trade-off relationship, we often use the equal error rate (EER), the point where $P_{Miss}$ and $P_{FalseAlarm}$ are equal, as a performance metric.
%
The NIST SRE also uses the same criteria. However, instead of EER, it employs a cost function $C_{\operatorname{Norm}}$ that weights one probability over the other. For example, it uses the following evaluation metric in the core test.
\begin{eqnarray}
    C_{\operatorname{Norm}} &=& P_{Miss}(\theta) + \beta \times P_{FalseAlarm}. 
\end{eqnarray}
$P_{Target}$, which is the prior probability that the target speaker is present in the speech segment to be matched, is lower than 0.5:
\begin{eqnarray}
    \beta &=& \frac{C_{FalseAlarm}}{C_{Miss}} \times \frac{1-P_{Target}}{P_{Target}},
\end{eqnarray}
%
where $C_{\operatorname{Miss}}$ is the unit cost of the false acceptance and $C_{\operatorname{FalseAlarm}}$ is the unit cost of the false rejection. 
In the NIST SRE2019 CTS (conversational telephone speech) test, $C_{\operatorname{Miss}}$ and $C_{\ \operatorname{FalseAlarm}}$ are set to 1 and $P_{\operatorname{Target}}$ is set to 0.01 or 0.005. This indicates that the evaluation is more concerned with false acceptances than with false rejections.

NIST SRE aims to evaluate one-to-one comparison performance. In this paper, we extend this to a one-to-many identification metric by incorporating an attack probability $\alpha$ and the unit cost $C$.



\section{Study Design}
We conducted a conjoint analysis\footnote{The entire study was approved by our Institutional Review Board.} to evaluate the impact of the implemented security measures of biometric system on attack occurrence probability $\alpha$. 

The objective of this study is to clarify the effect of the configuration change of the recognition system on the security risk.

\subsection{Participants}
To recruit eligible participants, we implemented a short screening survey prior to our main survey. The subjects of the questionnaire used in the conjoint analysis were 600 participants from 20 to 69 year-old living in Japan. In addition, Owing to the characteristics of the questionnaire, those without knowledge of biometrics were unsuitable for this survey. 
In particular, for the question "Please tell us about all recognition methods you know about biometrics," those who checked at least one of the options (face, palm, fingerprint, voice, or iris) were recruited in the main survey. All participants were required to complete consent forms before answering the main survey.

\subsection{Designing Conjoint Analysis}
\subsubsection*{Scenario}
\begin{figure}[t]
    \begin{center}
        \includegraphics[width=85mm]{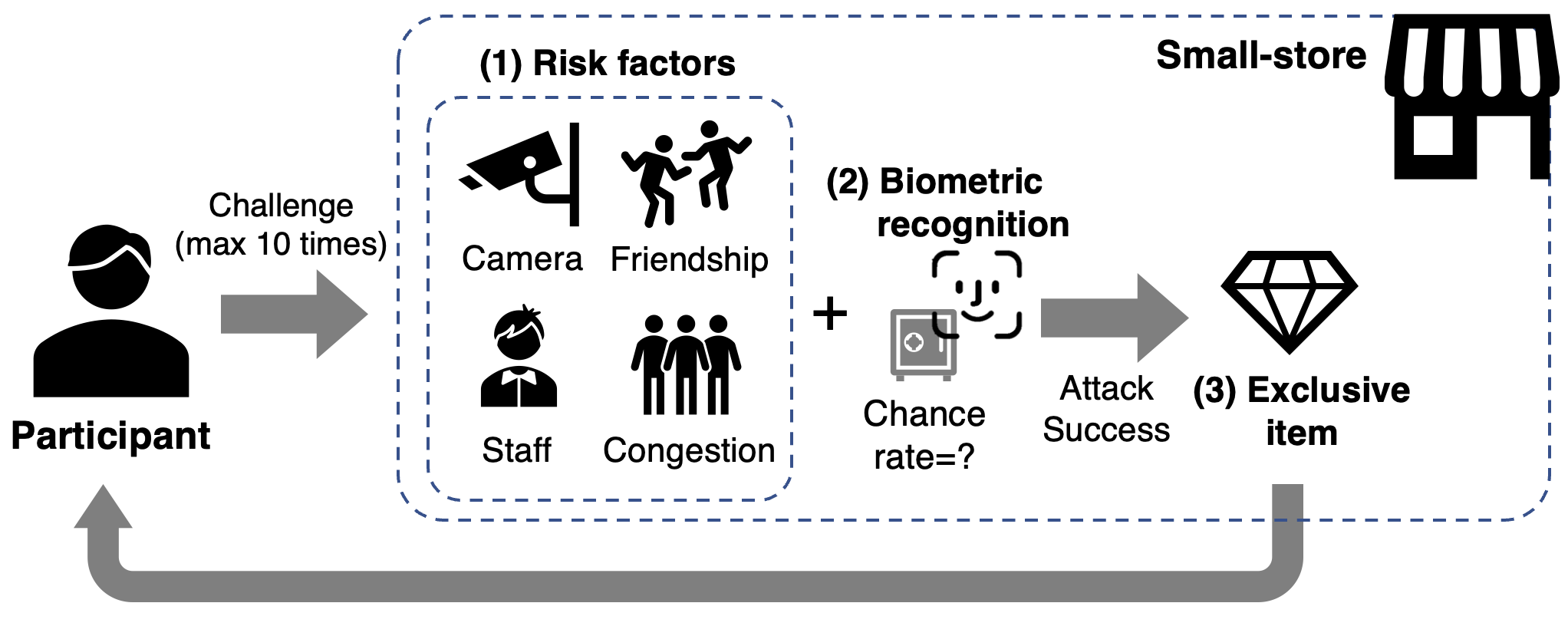}
        \caption{Overview of the study scenario. We planned a scenario in which the in-game challenge is to (1) break into a store with ``\textit{security measures}'' and (2) open a safe locked by ``\textit{biometric recognition}'' to (3) obtain a \textit{``exclusive item.''}.}
        \label{fig:scenario_overview}
    \end{center}
\end{figure}

In designing the conjoint analysis scenarios in this study, we considered scenarios in which theft in an unstaffed store would cause \textit{social-desirability bias} against committing the criminal act. Therefore, as shown in Fig. \ref{fig:scenario_overview}, we planned a scenario in which the in-game challenge is to
(1) break into a store with ``\textit{security measures}'' and (2) open a safe locked by ``\textit{biometric recognition}'' to (3) obtain a \textit{``exclusive item.''}.
We replaced the risk of arrest with a setting in which the shop allowed only a limited number of authorized persons, excluding participants.

When participants perform a game, they were allowed to attempt challenges ten times. The player can win the game by obtaining an exclusive item within ten attempts. However, if a surveillance camera or a store employee spots the user, the challenge is terminated midway. 

\begin{figure}[t]
    \begin{center}
        \includegraphics[width=85mm]{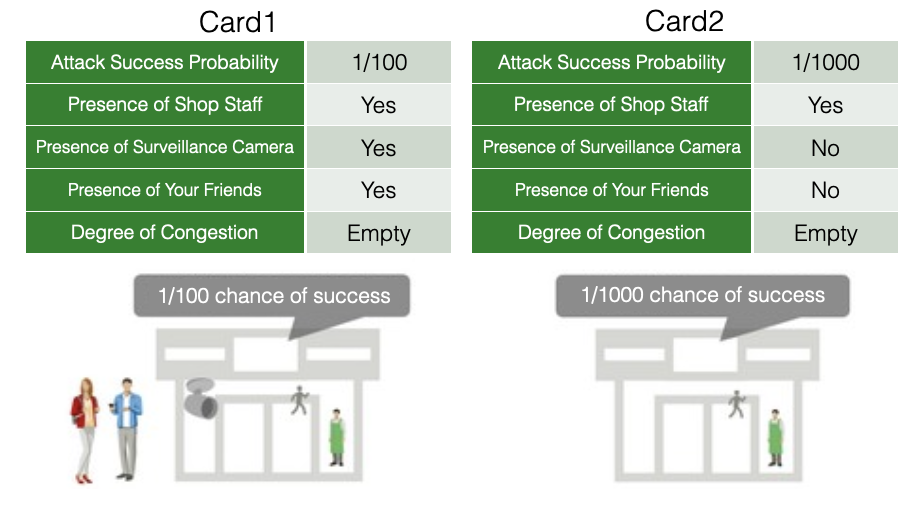}
        \caption{Example of pairwise comparison task: The actual task was conducted in Japanese, however, we show an English translated version for explanation}
        \label{fig:pair_question}
    \end{center}
\end{figure}

After explaining these conditions to the participants, we presented conjoint cards to the participants in a pairwise comparison method, as shown in Fig. \Ref{fig:pair_question}. Participants stated which of the recognition systems shown on the cards they thought to be more likely to win the game.

\subsubsection*{Attributes and Levels}
In this study, we selected five attributes, \texttt{FAR}, \texttt{Camera}, \texttt{Staff}, \texttt{Friendship} and \texttt{Congestion}. These were considered as risk factors against attackers in biometric recognition systems, particularly in small stores case.

We use the \texttt{FAR} condition for the performance of the recognition device, the \texttt{Camera} condition for the presence of a surveillance camera, and the \texttt{Staff} condition for the presence of shop staff.
In addition, we considered weak attackers who would attack out of mischievousness. These attackers may be strongly affected by psychological deterrents.
Therefore, we added a \texttt{Friendship} and \texttt{Congestion} as risk factors based on  psychological deterrents.
\texttt{Friendship} attribute represents a risk caused from the condition that \textit{``family members or friends may come to the store and meet by chance.''} The \texttt{Congestion} attribute represents a risk caused from a degree of congestion in the store with a 3-point scale (empty/normal/crowded). Table \ref{tab:conjoint_conditions} shows the attributes and levels used in this study.

\begin{table}[t]
    \centering
    \caption{Attributes and levels for conjoint analysis.}
    \begin{tabular}{cc}  \toprule
        Attributes &  Levels \\ \midrule

         \texttt{FAR} & $10^{-\{2,3,4,5\}}$ \\
         \texttt{Camera}   & Yes/No \\
         \texttt{Staff}     & Yes/No \\
         \texttt{Friendship}    & Yes/No \\
         \multirow{2}{*}{\texttt{Congestion}}  & empty/normal \\ 
                                                & /crowded       \\ \bottomrule
    \end{tabular}
    \label{tab:conjoint_conditions}
\end{table}

\subsubsection*{Conjoint Cards}\label{subsec:conjoint_card}
In this study, we conducted a conjoint analysis using a pairwise comparison method. The pairwise comparison method of analysis was defined such that participants were presented with two conjoint cards and asked to choose which they prefered. 
We created the conjoint cards using the following steps.
\begin{enumerate}
    \item Perform a full factorial design, which is a combination of all the levels of all attributes used. In the example in Table \ref{tab:conjoint_conditions}, $4 \times 2 \times 2 \times 2 \times 3 = 96$ possible combinations exist.
    \item Reduce the number of combinations because the number of combinations typically increases enormously with the number of attributes and levels in the full factorial design. This study used the optFederov() function included in AlgDesign, which is a package for designing experiments in R. The optFederov function takes the full factorial design and number of combinations to be reduced as arguments and reduces the number of combinations. Errors are likely to occur if the number of combinations is excessively large. The combination after reduction the previous steps is called the initial set.
    \item Create a copy of the initial and choice sets. Generate a random number corresponding to each row of the two initial sets, including the copy, and sort the initial sets in ascending order. Extract the attributes and levels from the same row of the two sorted initial sets. Consider the output as a conjoint card presented in a pair-wise comparison method. Note that the identical contents of the same row in the two initial sets output the same conjoint card, which makes questions meaningless. In this case, redo the procedure from the generation of random numbers.
\end{enumerate}
Following this procedure, we created nine conjoint cards, as shown in Table \ref{tab:conjoint_cards}. In each challenge, we present the conjoint cards in pairs in the order shown in Table \ref{tab:pair}.

\begin{table}[t]
\begin{minipage}[t]{.5\textwidth}
    \centering
    \caption{List of conjoint cards}
    \scalebox{0.75}{
        \begin{tabular}{cccccc}  \toprule
            Index &  \texttt{FAR} & \texttt{Camera}	& \texttt{Staff} & \texttt{Relationship} & \texttt{Congestion} \\ \midrule
            1   &  $10^{-2}$	& Yes	& Yes   & Yes   & empty \\
            2   &  $10^{-2}$	& No	& No    & No    & normal \\
            3   &  $10^{-3}$	& Yes	& No    & Yes   & crowded \\
            4   &  $10^{-3}$	& Yes	& Yes   & Yes   & normal \\
            5   &  $10^{-3}$    & No	& Yes   & No    & empty \\
            6   &  $10^{-4}$	& Yes	& No    & No    & empty \\
            7   &  $10^{-4}$    & No	& Yes   & Yes   & normal \\
            8   &  $10^{-5}$	& No	& No    & Yes   & empty \\
            9   &  $10^{-5}$    & Yes   & Yes   & No    & crowded \\ \bottomrule
        \end{tabular}
    }
\label{tab:conjoint_cards}
\end{minipage}

\vspace{3mm}
\begin{minipage}[t]{.45\textwidth}
\centering
    \caption{Pairwise comparison table}
    \begin{tabular}{ccc}
     \toprule
        Number	& Card1	& Card2 \\ \midrule
        1	& 1	    & 5 \\
        2	& 9	    & 7 \\
        3	& 8	    & 1 \\
        4	& 5	    & 4 \\
        5	& 6	    & 2 \\
        6	& 7	    & 8 \\
        7	& 4	    & 3 \\
        8	& 3	    & 6 \\
        9	& 2	    & 9 \\ \bottomrule
    \end{tabular}
    \label{tab:pair}
\end{minipage}    
\end{table}

\subsubsection*{Analyze}
To analyze which attributes contributed to the choice of conjoint cards, we applied conditional logistic regression to the survey results. The objective variable in the logistic regression equation was the selected conjoint card, and the explanatory variables were all attributes. We performed conditional logistic regression analysis using the clogit() function included in survival, which is a survival analysis package in R. The clogit calculates the coefficient estimates (coef), odds ratio (exp), standard error of the exp (se), z-value (z), and p-value (p) for each attribute using the objective and explanatory variable as input. This allowed us to analyze the impact of each attribute on the choice of conjoint cards. 
An increase in the coefficient estimate means that the attribute contributed significantly to the conjoint card selection. In contrast, a decrease in the coefficient estimate means that the attribute did not contribute to the conjoint card selection.

\subsection{Risk Evaluation}

\subsubsection*{Metrics}
We define a risk evaluation metric for the identification system to evaluate the risk considering the probability of an attacker and the amount of damage caused by an attack based on standard NIST SRE metrics \cite{NISTSRE}. 

In an identification system, both the risk of impersonation by non-registered users and that of registered users must be considered.
First, we define a value $\operatorname{P_{open}}$ that evaluates the impersonation probability of non-registered users. Let $\operatorname{P_{open}}$ be the probability of a False Accept in identification trials after an exhaustive search through a database of $N$ unrelated templates. Let $\operatorname{P_{FA}}$ be the probability of a False Accept in a verification trial. Daugman \cite{daugman2000biometric} defines $\operatorname{P_{open}}$ as False Accept among those $N$ comparisons is one minus that probability. 
\begin{eqnarray}
\operatorname{P_{open}}=(1 - ( 1 - \operatorname{P_{FA}})^N)
\label{eq:fpiropen}
\end{eqnarray}

Next, we define $\operatorname{FPIR_{close}}$, which evaluates the impersonation probability of registered users. $\operatorname{FPIR_{close}}$ can be defined as the probability of a false match occurring with any registered user other than the user attempting recognition. Let $\operatorname{P_{FR}}$ be the probability of a False Reject in a verification trial. Considering Equation (\ref{eq:fpiropen}), $\operatorname{P_{close}}$ can be defined as follows:
\begin{eqnarray}
\operatorname{P_{close}}=\operatorname{P_{FR}}(1 - ( 1 - \operatorname{P_{FA}})^{N-1})
\label{eq:fpirclose}
\end{eqnarray}
As described in \cite{daugman2000biometric}, $\operatorname{FPIR_{open}}, \operatorname{FPIR_{close}}$ can be approximated by $\operatorname{FPIR_{open}} \approx NP_{FA}$ and $\operatorname{FPIR_{close}} \approx \operatorname{P}_{FR}(N-1)\operatorname{P}_{FA}$ for small $\operatorname{P}_{FA}$ or small $\operatorname{P}_{FR} \ll \frac{1}{N} \ll 1$. When searching a database of size $N$, an identifier needs to be roughly N times better than a verifier to achieve comparable odds against a False Accept.

Considering $\operatorname{FPIR_{open}}$ and $\operatorname{FPIR_{close}}$ and let $\alpha$ be the probability of all recognition transaction occurrences by non-registered users, i.e., by malicious attackers, and let $C$ be the assumed damage cost. Using these, we define the risk value evaluation metric $\operatorname{C}_{identify}$ by the following equation:
%
%
\begin{align}\label{eq:cidentify}
 \operatorname{C}_{identify} =
 & C_{open} \cdot \alpha \operatorname{FPIR_{open}}  \nonumber \\ 
 & + C_{close} \cdot (1-\alpha) \operatorname{FPIR_{close}}.
\end{align}
$\mathcal{C}_{identify}$ enables the risk analysis of biometrics considering the attack occurrence probability $\alpha$. Suppose identifying the impact of various environmental factors on the probability of an attack is possible. This allows for the comparison of risks between different environments and use cases using an index that is the product of the number of false acceptances that may occur within a unit period and the cost of damage caused by false acceptances. This can serve as a reference for system design when introducing a new recognition system.

\label{sec:evaluate_cidentify}
The $\alpha$ for a particular attribute and level combination can be calculated using the perceived values for each attribute calculated in the conjoint analysis as follows:
\begin{eqnarray}
 \alpha &=& \sum_{a \in \mathcal{A}}l_a/l^* \label{eq:alpha}, \\
 l^* &=& \sum_{a \in \mathcal{A}}|l_a|, \\
 C_{open} &=& C_{close} = 0.5, \label{eq:copencclose}
\end{eqnarray}
where $\mathcal{A}$ is the set of attributes ($\{\texttt{FAR}, \texttt{Camera}, \cdots\}$), $|l_a|$ the number of levels of attributes $a$, and $l_a=[0, |l_a|-1], l_a \in \mathbb{Z}$ the level selected for attributes $a$ with an integer step value. 
Note that $l^*$ is used to normalize $\alpha$ such that $\alpha=1.0$ when all security measures are not applied (weakest security measure) and $\alpha=0.0$ when all security measures are applied at their strongest settings.
As shown in Table \ref{tab:conjoint_conditions}, we set the deterrence such that the larger the $l_a$, the higher the deterrence for all levels. 
For the cost in Equation (\ref{eq:copencclose}), because this experiment assumes a small store, we assumed that a significant difference does not exist between $C_{open}$ and $C_{close}$ in either case.

\section{Results}\label{sec:result}
%
\begin{table*}[t]
    \centering
    \caption{Result of the conjoint analysis ($^{*}$: \textbf{p$<$0.1}, $^{**}$: \textbf{p$<$0.05})}
    \begin{tabular}{cccccc} \toprule
                            & coef     	& exp (coef) & se (coef) & z 	     & p \\ \midrule
        \texttt{FAR}      	        & -0.460	& 0.632	    & 0.022	   & -21.074 & \textless\textbf{}{2e-16$^{**}$} \\
        \texttt{Staff}	& -0.093	& 0.911     & 0.052	   & -1.79	 & $0.073^*$ \\
        \texttt{Camera}	        & -0.336	& 0.715	    & 0.041	   & -8.119 & \textless\textbf{2e-16$^{**}$} \\ \texttt{Friendship}	        & -0.056	& 0.946	    & 0.052	   & -1.085 & 0.278 \\ \texttt{Congestion}	        & -0.169	& 0.845	    & 0.028	   & -5.978 & \textless\textbf{2e-16$^{**}$} \\ \bottomrule        
    \end{tabular}
    \label{tab:conjoint_result}
\end{table*}
\subsection{Conjoint analysis}
Table \ref{tab:conjoint_result} shows the evaluation of the survey results by conjoint analysis.

In particular, this discussion focuses on the values in the \textit{coef} column because they represent the utility values, which indicate the importance of different attributes in the survey. 
The \textit{coef} column shows that the negative factors due to \texttt{FAR}, \texttt{Camera}, and \texttt{Congestion} are significant (p$<$0.05) with values of -0.460, -0.336, and -0.169, respectively. \texttt{Staff} has a marginal significance (p$<$0.1) at -0.093, while \texttt{Friendship} is insignificant at -0.056.
These results indicate that deterrents like \texttt{FAR}, \texttt{Camera}, and \texttt{Congestion} were considered much more effective than factors such as \texttt{Staff}, or \texttt{Friendship}.
Furthermore, the combined effect of \texttt{Camera} and \texttt{Congestion} is smaller than that of \texttt{FAR}. This indicates that using surveillance cameras in crowded stores is almost as effective as tightening the \texttt{FAR} (e.g., changing from FAR=$10^{-3}$ to FAR=$10^{-4}$)

\subsection{Use cases}
We verify the effectiveness of conjoint analysis by comparing the value of $\mathcal{C}_{identify}$ across typical use cases. 
Specifically, we assume three use cases: (1) Low-security with no measures, (2) Mid-security with some measures, and (3) High-security with all measures.
Table \ref{tab:cidentify_attrs_and_levels} lists the attributes and levels for each use case. In each case, we compare the value of $\mathcal{C}_{identify}$ with the change in FAR. 
Note that although the p-value for \texttt{Friendship} was insignificantly different, this study uses the value of \texttt{Friendship} in the calculation of $C_{identity}$. 
This is because the perceived value of \texttt{Friendship} is minimal, and its overall impact is negligible.

Table \ref{tab:cidentify_result1} shows the $\mathcal{C}_{identify}$ values for each use case and FAR combination (FRR=$10^{-2}$, N=10000).

The reference value is the $\mathcal{C}_{identify}$ value of 0.293, shown in the light-grey cell under the Low-secure column at FAR=$10^{-4}$.
Dark-gray cells indicate conditions where $\mathcal{C}_{identify}$ is lower than the reference value.
We observe that in the High-secure case at FAR=$10^{-3}$, the $\mathcal{C}_{identify}$ value is lower than the reference value, even though the FAR is one step lower.
This result indicates that a recognition algorithm with a higher FAR can achieve the same level of security as one with a lower FAR by applying high-security measures.

\begin{table}[t]
    \centering
    \caption{Attributes and levels for each use case. (1) Low-secure: no security measures are taken, (2) Mid-secure: some security measures are taken, and (3) High-secure: all security measures are taken}
    \begin{tabular}{cccc}
    \toprule
        & \multicolumn{3}{c}{Use cases} \\ \cmidrule{2-4}
        & Low-secure       & Mid-secure       & High-secure \\ \midrule
\texttt{Staff}         & No       & Yes        & Yes \\
\texttt{Camera}        & No       & No       & Yes \\
\texttt{Friendship}    & No       & Yes        & Yes \\
\texttt{Congestion}    & Empty   & Normal  & Crowded \\ \bottomrule
    \end{tabular}
\label{tab:cidentify_attrs_and_levels}
\end{table}
\begin{table}[t]
    \centering
    \caption{$\mathcal{C}_{identify}$ values for each use case and FAR combinations (FRR=$10^{-2}$, N=10000): Dark grey cell denotes a combination where $\mathcal{C}_{identify}$ is smaller than in the cell of row Low-secure and column FAR=$10^{-4}$.}
        \begin{tabular}{ccccc}
    \toprule
\multicolumn{2}{c}{} & \multicolumn{3}{c}{Use cases} \\ \cmidrule{3-5}
\multicolumn{2}{c}{} & Low-secure  	& Mid-secure  	& High-secure \\ \midrule
\multirow{4}{*}{\texttt{FAR}} & $10^{-2}$	    & 0.5   & 0.390	& 0.315 \\
& $10^{-3}$	    & 0.406	        & 0.296	& \cellcolor[rgb]{0.7, 0.7, 0.7}0.211 \\
& $10^{-4}$	    & \cellcolor[rgb]{0.9, 0.9, 0.9}0.293	        & \cellcolor[rgb]{0.7, 0.7, 0.7} 0.127	& \cellcolor[rgb]{0.7, 0.7, 0.7}0.108 \\
& $10^{-5}$    &  \cellcolor[rgb]{0.7, 0.7, 0.7} 0.019	& \cellcolor[rgb]{0.7, 0.7, 0.7} 0.010	& \cellcolor[rgb]{0.7, 0.7, 0.7} 4.99e-4 \\ \bottomrule
    \end{tabular}
\label{tab:cidentify_result1}
\end{table}
\section{Discussion and Limitation}
\subsection{Discussion}
As shown in Table \ref{tab:conjoint_result}, the negative impact of factors, such as \texttt{FAR} and \texttt{Camera}, is significant. Additionally, from Table \ref{tab:cidentify_result1}, $\mathcal{C}_{identify}$ allows us to compare use cases that consider various risk factors. This section discusses some of the factors that can affect the evaluation results.


\subsubsection*{Impact of FAR}
Table \ref{tab:cidentify_result1} show the usefulness of $\mathcal{C}_{identify}$ by comparing \texttt{FAR} with various use cases. We believe that $\mathcal{C}_{identify}$ is useful for system design analysis when introducing a new biometric system with \texttt{FAR}, considering additional measures from existing systems.
In this study, \texttt{FAR} was graded using powers of 10 and presented to the participants. However, a proportional relationship does not necessarily exist  between the small \texttt{FAR} and the probability of attack occurrences. Instead, a relationship may exist such that the probability of attacks decreases significantly after a particular value.

Therefore, more detailed user tendencies might need to be considered in the evaluation. For example, for some perceived values, it may be necessary to consider a modeling method such as a sigmoid function instead of Equation (\ref{eq:alpha}), may be necessary.

\subsubsection*{Impact of rewards}
In this study, participants played a game to obtain an unspecified exclusive item. However, in real scenarios, rewards depend on the target system (e.g., a jewelry store versus a small-scale store). In the pilot study, specifying a reward amount led participants to base actions solely on the reward. Therefore, we redesigned the scenarios to analyze the impact of other factors on attack probability without specifying reward amounts.

\subsection{Limitation}
\subsubsection*{Social desirability bias}
To eliminate social desirability bias, this study was conducted under the scenario of a game in which players tried to obtain an exclusive item. Therefore, the $\mathcal{C}_{identify}$ obtained in this study may differ from the $\mathcal{C}_{identify}$ for biometric systems under realistic conditions. In addition to the interrelationships of $\mathcal{C}_{identify}$ with the varying risk factors identified in this study, future work must examine how best to evaluate $\mathcal{C}_{identify}$ under realistic scenarios.

\subsubsection*{Attack by registered users}
In this study, we evaluated $\mathcal{C}_{identify}$ by considering registered users as attackers and non-registered users as non-attackers. While a registered user can attack in a real system, which may affect the risk evaluation results, this scenario is less likely to occur. Therefore, we did not consider it in this study.

\subsubsection*{Considerations for practical implementation}
In this study, we proposed and validated the framework employing typical use cases. However, in practice, other factors specific to the implementation environment and other use cases may also need to be considered. These considerations are future challenges to be addressed.

\section{Conclusion}
In this study, we proposed a novel human-centered risk evaluation framework for biometric systems using conjoint analysis. Our framework focuses on quantifying the impact of various risk factors on attackers' motivation, enabling a comprehensive risk assessment that goes beyond traditional metrics such as the False Acceptance Rate (FAR). Through a survey of 600 Japanese participants, we demonstrated how different security measures, including the presence of surveillance cameras and store staff, can significantly influence the probability of attacks. The findings highlight the importance of considering human factors in the design and implementation of biometric systems to enhance security without compromising usability. 
%
This approach is expected to facilitate the development of more reliable and secure biometric systems. Future research will expand on these insights by conducting larger-scale studies and applying the framework in practical scenarios to further refine our understanding of biometric system risks and their mitigation.
\section*{Acknowledgement}
This work was supported in part by JST Moonshot R\&D Grant Number JPMJMS2215.

\bibliographystyle{ieee}
\bibliography{references}

\end{document}